\documentclass[10pt,twocolumn,letterpaper]{article}

\usepackage{cvm}
\usepackage{times}
\usepackage{epsfig}
\usepackage{graphicx}
\usepackage{amsmath}
\usepackage{amssymb}
\usepackage{authblk} 
\usepackage[ruled,vlined]{algorithm2e}
\usepackage{caption}
\usepackage{booktabs}
\usepackage{multirow}
\usepackage{pifont}

\usepackage[pagebackref=true,breaklinks=true,letterpaper=true,colorlinks,bookmarks=false]{hyperref}
\usepackage[capitalize]{cleveref}
\usepackage[capitalize]{cleveref}
\crefname{section}{Sec.}{Secs.}
\Crefname{section}{Section}{Sections}
\Crefname{table}{Table}{Tables}
\crefname{table}{Table.}{Tables.}
\crefname{figure}{Figure.}{Figures.}

\newcommand{\keywords}[1]{{\bf \emph{Keywords: #1}}}

\cvmfinalcopy

\ifcvmfinal\fi
\begin{document}

\title{Unwarping Screen Content Images via Structure-texture Enhancement Network and Transformation Self-estimation}
\author{Zhenzhen Xiao, Heng Liu*, and Bingwen Hu}
\maketitle
\begin{abstract}
    While existing implicit neural network-based image unwarping methods perform well on natural images, they struggle to handle screen content images (SCIs), which often contain large geometric distortions, text, symbols, and sharp edges. To address this, we propose a structure-texture enhancement network (STEN) with transformation self-estimation for SCI warping. STEN integrates a B-spline implicit neural representation module and a transformation error estimation and self-correction algorithm. It comprises two branches: the structure estimation branch (SEB), which enhances local aggregation and global dependency modeling, and the texture estimation branch (TEB), which improves texture detail synthesis using B-spline implicit neural representation. Additionally, the transformation self-estimation module autonomously estimates the transformation error and corrects the coordinate transformation matrix, effectively handling real-world image distortions. Extensive experiments on public SCI datasets demonstrate that our approach significantly outperforms state-of-the-art methods. Comparisons on well-known natural image datasets also show the potential of our approach for natural image distortion.
\end{abstract}


\section{Introduction}

Image warping \cite{son2021srwarp,lee2022learning} is a vital image processing technique that alters the spatial positions of pixels to achieve specific visual effects or changes in image morphology. This process involves adjusting the coordinates of each pixel in the image and mapping them to the new positions, thereby changing the shape or perspective of the image. In practice, image warping is often combined with other image processing techniques, such as image correction~\cite{chiang2000efficient,chiang1996efficient},repositioning~\cite{sarlin2021back,von2020lm}, and lens distortion correction~\cite{aleman2014automatic,santana2016iterative,li2017robust}, to achieve comprehensive visual enhancement. The technique relies on mathematical models and algorithms to compute the necessary pixel mappings, making it a versatile and crucial tool in visual effects processing and real-world image correction.

Image warping aims to deform images defined on a rectangular grid into continuous shapes. By using the precise transformation or mapping function, one can effectively reconstruct the spatial structure of an image with an arbitrary shape and scale to achieve the desired image transformation effect. SRwarp~\cite{son2021srwarp} first redefines the image warping problem as a spatial variation of generalized super-resolution and effectively addresses the reconstruction of high-frequency detail images by utilizing a deep single-image super-resolution (SISR)~\cite {yang2019deep, dai2019second,lim2017enhanced}architecture as the backbone.

\vspace{+2em}
\begin{figure}[t]
    \centering
    \begin{minipage}[t]{0.11\textwidth} 
        \centering     
        \includegraphics[width=\textwidth]{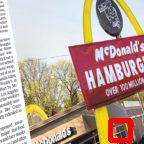}
    \centerline{Warped Image} 
    \end{minipage}
    \hspace{0.1cm} 
    \begin{minipage}[t]{0.11\textwidth}
        \centering
        \includegraphics[width=\textwidth]{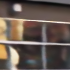}
        \vspace{\baselineskip}
        \centerline{SRWarp\cite{son2021srwarp}} 
    \end{minipage}
    \begin{minipage}[t]{0.11\textwidth}
        \centering
        \includegraphics[width=\textwidth]{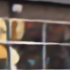}
        \vspace{\baselineskip}
        \centerline{LTEW\cite{lee2022learning}}
    \end{minipage}
     \begin{minipage}[t]{0.11\textwidth}
        \centering
        \includegraphics[width=\textwidth]{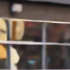}
         \vspace{\baselineskip}
         \centerline{STEN(Ours)} 
    \end{minipage}
    \caption{Real-world screen content image correction employs the proposed STEN method. The image is captured by a camera, and both SRWarp\cite{son2021srwarp} and LTEW\cite{lee2022learning}, along with our proposed STEN, utilize the predicted transformation matrix to achieve super-resolution (SR).
    }
    \label{fig1}
\end{figure}

In recent years, inspired by implicit neural functions ~\cite{sitzmann2020implicit,lee2022local,pak2023b}, implicit neural representations have shown significant potential and effectiveness in handling image distortions. LIIF~\cite{chen2021learning} parameterizes the implicit neural representation using multi-layer perceptrons (MLP), mapping the input coordinates and corresponding latent variables to RGB values, thereby achieving super-resolution reconstruction at arbitrary scales. However, due to a lack of spatial variation considerations, LIIF~\cite{chen2021learning} has limitations in addressing image warping. Although the above methods can achieve rather good effects in dealing with natural image distortions, they often lose their effectiveness when facing Screen Content Images(SCIs) with huge deformations, including geometric shapes, text, symbols, and sharp edges. Given this, LTEW~\cite{lee2022learning} utilizes the Fourier feature representations and the Jacobian matrices of coordinate transformation to continuously transform the input SCI into various shaped images. However, the signal representation obtained from LTEW~\cite{lee2022learning} only includes the finite sinusoidal components, which can lead to the reconstructed values revealing the Gibbs phenomenon\cite{gottlieb1997gibbs,gottlieb2011review}. Ensuring the excellent reconstruction of the specific image structures and textures at any scale and shape is crucial for SCIs unwarping.

In this work, we propose a structure-texture enhancement model with transformation self-estimation for SCIs warping. Our model employs an implicit neural representation of the B-spline texture coefficient estimator~\cite{pak2023b} of SCIs to mitigate the risk of undershoots and overshoots during reconstruction. Our model consists of two main branches: the structure estimation branch and the texture estimation branch. The structure estimation branch employs the Transformer technology\cite{liang2021swinir,zheng2024efficient,ashish2017attention}, introducing global attention and local attention modules that operate in parallel to enhance structure features, thus facilitating effective global modelling.
The texture estimation branch utilizes the B-spline features estimated from the input images and the Jacobian matrix of coordinate transformations. In geometry, the determinant of the Jacobian matrix \cite{umetani1989resolved,chen2017tracking}represents local magnification ratios, thereby enhancing the ability of the network to learn texture information. Inspired by LTEW \cite{lee2022learning}, the Jacobian matrix of spatial changes is multiplied by the B-spline features of each pixel to compute pixel shapes described by directional curvature, further enhancing the ability of the network to learn texture information.
Since structural features better represent the overall shape and structure of SCI images, while texture features capture details and texture characteristics, we propose a structure-texture fusion module to enhance the representation capability of image features, making them more comprehensive and diverse.

In addition, we design a transformation self-estimation algorithm based on training a transformation error estimation network to obtain the correct transformation matrix in real-world uncalibrated scenarios. As shown in Fig.~\ref{fig1}, based on the warped input image, our method can automatically and correctly estimate the transformation matrix. Even with the same transformation matrix, our approach can generate clearer, unwarped results than existing methods.

In summary, our contributions can be outlined as follows:
\begin{itemize}
\item We propose a dual-branch structure-texture enhancement model for SCIs warping. Our approach enhances image texture details through feature-based structure and texture enhancements.
\item We introduce a transformation self-estimation algorithm based on training a transformation error estimation network that can predict the correct coordinate transformation matrix for the corresponding warped image, thereby enhancing the robustness of our model in real-world scenarios.
\item Comprehensive comparative experiments and ablation studies demonstrate the superior performance and effectiveness of our unwarping approach across three SRC datasets and five natural image datasets, achieving state-of-the-art results for various scaling factors. 
\end{itemize}

\section{Related Works}
\subsection{Image Warping}
Image warping refers to the process of changing or reshaping the shape or visual appearance of an image by remapping or transforming pixels within it. This technique is widely used in computer vision\cite{voulodimos2018deep}, computer graphics\cite{agoston2005computer}, and image processing\cite{van2014scikit} for various purposes, including image correction, enhancement, and special effects. One common approach is grid transformation~\cite{zhou2005large,mok2020large}, which involves defining a grid within the image and locally deforming pixels within this grid to achieve more precise shape adjustments and deformation effects. Through warping, the shape and size of objects in an image can be modified to enhance their visual quality and appearance. SRWarp~\cite{son2021srwarp} interprets the image warping task as a spatial transformation problem within the super-resolution framework, proposing a method for handling arbitrary image transformations. However, the generalization capability of SRWarp is limited when dealing with unseen transformations, such as homography transformations with significant magnification factors. LTEW~\cite{lee2022learning} builds on the advantages of Fourier features and the spatial variation Jacobian matrix of coordinate transformations, introducing a continuous neural representation for image distortion. Despite its advancements, LTEW's use of Fourier representation, with its finite sinusoidal components, can lead to signal undershoots or overshoots at discontinuities, a phenomenon known as the Gibbs phenomenon~\cite{gottlieb1997gibbs,gottlieb2011review}.

When reconstructing signals with discontinuities \cite{duffin1952class} (such as step changes), the Gibbs phenomenon is prone to occur. Obviously, for SCIs, due to the presence of text, shapes, and symbols with large jumps or sharp edges, it is vital to overcome the Gibbs phenomenon and achieve high-definition reconstruction when performing unwarping transformations on them.
\vspace{-0.5em}
\subsection{Implicit neural representation (INR)}
Implicit Neural Representations (INR) \cite{chen2021learning,lee2022local,sitzmann2020implicit,pak2023b} utilize neural networks to provide implicit and continuous representations of signals, demonstrating significant advantages in handling spatial resolution and arbitrary coordinate transformations. The core idea of INR is to learn an implicit representation of a signal through neural networks, which is defined in a continuous space, thus overcoming the limitations of traditional grid-based representation methods. Specifically, implicit neural representations can finely describe signals, effectively handling and reconstructing them even under high resolutions and complex transformations. The LTEW \cite{lee2022learning} method combines local texture estimators with feature maps from deep neural network encoders and relative coordinates (or local grids), using local implicit neural representations (INR) to enhance the spatial resolution of input signals. This method performs exceptionally well during training and generalizes effectively to unseen tasks, showcasing robustness and versatility in complex environments and transformations. This paper proposes a new approach that combines B-spline basis functions \cite{pak2023b,prasad2022nurbs} with relative coordinates or local grids to more effectively handle high discontinuities in signals, such as Screen Content Images (SCIs). By incorporating B-spline basis functions, this method improves accuracy at discontinuities, thereby further extending the potential and effectiveness of implicit neural representations in practical applications.
\vspace{-0.5em}
\subsection{B-spline representation}
B-spline is a mathematical tool widely used in computer graphics\cite{foley1996computer,li2020application}, image processing\cite{lehmann2001addendum,chitra2021retracted}, and signal processing\cite{schoenberg1946contributions,unser1993b,unser1993b1}. Its representation consists of control points, basis functions, and a knot vector. Control points define the shape of the curve, while basis functions are used to combine these control points, and the knot vector determines the support intervals of the basis functions. Unlike traditional methods that rely on signal resolution, B-spline representation achieves efficient memory management through linear scaling of model parameters. In recent years, it has gained significant attention in the field of signal processing\cite{unser1993b}, particularly for handling highly discontinuous images, such as SCI, where B-spline methods excel in accurately reconstructing local features and image shapes.

In current research, implicit B-spline representation is widely utilized for distortion removal and reconstruction. Researchers have increasingly combined trainable B-spline basis functions with Non-Uniform Rational B-Splines (NURBS) layers\cite{prasad2022nurbs,pak2023b}, achieving notable success in geometric modeling and complex signal reconstruction. These methods can accurately estimate coefficient information and extract structural and distortion information from local regions and advanced features of input images. Building on these research achievements, we integrate deep neural network backbones with B-spline basis functions to represent deformed images under arbitrary coordinate transformations, further enhancing processing effectiveness.

\begin{figure*}[t]
  \centering
   \includegraphics[width=1\linewidth]{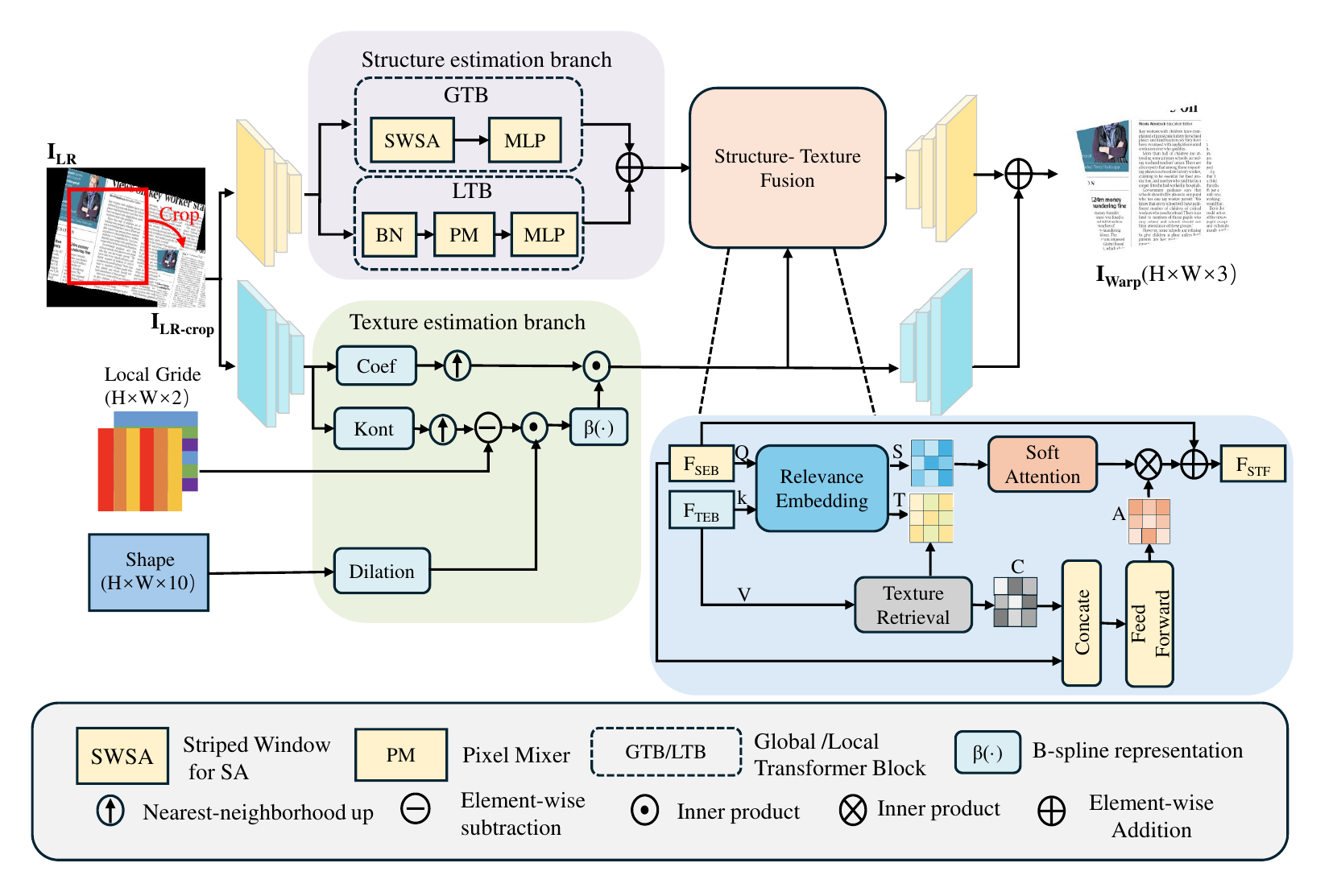}
   \caption{The overall structure of the proposed STEN model is as follows: First, the low-resolution samples \( I_{LR} \) are subjected to an inverse transformation using the inverse transformation matrix \( M^{-1} \), followed by cropping to obtain \( I_{LR-\text{Crop}} \) images. The entire model consists of two main branches: the Structure branch and the texture branch. The structure estimation branch allows us to obtain richer structural features, while the texture estimation branch enhances the ability of the proposed method to estimate texture information and local details more effectively, thus extracting texture features more efficiently. These features from both branches are then mapped into the Structure-Texture Fusion (STF) module for feature enhancement and subsequently decoded. By combining the Structure and texture branches, our STEN model can better handle deformation issues in screen content images.
   }
   \label{fig2}
   \vspace{-1em}
\end{figure*}

\section{Method}
In this section, our objective is to reconstruct a high-resolution warped image ${I^{WARP} \in {R}^{3\times H\times W}}$ from a low-resolution RGB image ${I_{LR} \in {R}^{3\times h\times w}}$ using a differentiable and invertible coordinate transformation ${\textit{f}: X \rightarrow Y}$. Here, $X$ represents the input coordinate space, which is a set of points on a 2D plane, and  $Y$ denotes the output coordinate space after transformation. To prepare the input image $I_{LR}$, we first apply the inverse coordinate transformation:
\begin{equation}
  {I_{LR} =I_{HR}[\textit{f}^{-1} (Y)]}.
  \label{eq:important1}
\end{equation}
In practice, warping modifies the image resolution while preserving pixel density, transforming it from (\( h \times w \rightarrow H \times W \)).

In traditional mathematical methods, addressing image warping issues often relies on point-matching techniques, which are widely used in multi-view stereo reconstruction. By precisely matching corresponding points from different views, the geometric transformations of images can be derived. However, this point-matching method exhibits significant limitations when handling large-scale or complex deformation scenarios. In such cases, extracting and matching feature points becomes challenging, leading to poor performance in large-range warping tasks. In recent years, end-to-end deep learning models with implicit Fourier neural representation \cite{son2021srwarp,lee2022learning,xiao2024towards} have been introduced to tackle unknown image transformations. However, these methods still face challenges when dealing with complex and sharp geometric transformations. This is because the Fourier representation typically relies on a finite set of sine waves for local signal analysis and reconstruction. While it performs well in processing smooth frequency variations, it is prone to Gibbs phenomena when dealing with SCIs that exhibit discontinuous texture features, causing oscillations at the edges of images.

In contrast, B-spline representation utilizes piecewise polynomials for local interpolation, which has significant advantages in capturing local texture details and high-frequency components of images. In addition to fine texture reconstruction, it is also necessary to consider the structural integrity of SCIs when unwarping. Due to the powerful contextual information retention capability of the Transformer, both global and local Transformers are used to maintain the structural integrity of SCIs. Thus, in this work, we propose a structure-texture enhancement network (STEN) for SCI warping, which combines implicit B-spline neural representation for texture estimation and global-local transformers for structure estimation.

\subsection{Overview}
The overall architecture of our proposed STEN is illustrated in Fig.~\ref{fig2}.
Our model consists of two main branches, \emph{i.e.,} the structure estimation branch (SEB) and the texture estimation branch (TEB). Each incorporates an encoder and a decoder. Initially, we extract the features $F_{LR}$ from the low-resolution input image using the encoder and then feed them into the Structure estimation branch and the texture estimation branch of the model.

In the Structure estimation branch, we enhance the aggregation of local knowledge through channel shift and mix operations (called pixel mixer, PM) and utilize image anisotropy for effective global dependency modelling, thereby obtaining richer structural features. Meanwhile, in the B-spline texture estimation branch, we improve the capability of the network to estimate texture information and local details, thereby enhancing texture features. Given that SCIs often have discontinuous tones and many similar text regions, we design a structure-texture fusion module to globally retrieve texture features and integrate the most relevant information into the structural features for feature-based texture enhancement. This module combines the image's structural information with texture features, effectively enhancing image details and visual effects while preserving the structural integrity of the image, resulting in clearer and more vivid images. 
Finally, the fused features and texture features are reconstructed through decoders separately. The decoders are constructed with multi-layer perceptrons (MLPs) and predict the RGB values at queried coordinates. The predicted values are then added to produce the final reconstructed high-resolution un-warped image.

\subsection{Structure Estimation Branch}
As shown in Fig.~\ref{fig2}, the Structure estimation branch consists of two types of transformer modules: the Local Transformer Block (LTB) and the Global Transformer Block (GTB), which process the features \( F_{SLR} \) from the encoder. 

In the Local Transformer Block (LTB), we first use a Batch Normalization (BN) layer to stabilize and accelerate training, and then use a Pixel Mixer (PM) layer to enhance the aggregation of local information. Specifically, PM divides the feature channels into four equally sized groups, and each group undergoes a specific shift sequence (left, right, top, bottom) to introduce locality and spatial correlations. This approach allows us to quickly capture and integrate knowledge from surrounding areas while improving the effective information exchange mechanisms between feature channels. The entire process of the Local Transformer Block (LTB) can be expressed as:
\begin{equation}
  F_{L} = MLP(PM(BM(F_{LR}^{S}) )),
  \label{eq:important2}
\end{equation}
where PM denotes the Pixel Mixer, BM represents Batch Normalization, and MLP refers to the multi-layer perceptron used for further feature transformation, \( F_{LR}^{S} \) being the input encoded features of the structure estimation branch.

In the Global Transformer Block (GTB), we construct an effective global dependency model based on self-attention (SA). We employ stripe window self-attention (SWSA) to effectively capture multi-scale symmetries and similarities present in the image. Firstly, self-attention is computed for each stripe window:

\begin{equation}
   F_{out}^n = \text{Softmax}\left(\frac{ Q^n \cdot (Q^n)^T}{\text{scale}}\right) \cdot V^n,
  \label{eq:also-important1}
\end{equation}
where \( n \) denotes the \( n \)-th stripe window, with \( Q^n = F_{LR}^{S} \cdot W_Q \) and \( V^n = F_{LR}^{S} \cdot W_V \) being linear transformations of $F_{LR}^{S}$ using weight matrices \( W_Q \) and \( W_V \). Subsequently, the outputs of all stripe windows are concatenated along the channel dimension:

\begin{equation}
  F_{out} = [F_{out}^1 ; F_{out}^2 ; \ldots ; F_{out}^n].
  \label{eq:also-important2}
\end{equation}
Next, \( F_{out} \) undergoes further feature transformation through a multi-layer perceptron (MLP):
\begin{equation}
  F_{G} = MLP(F_{out}).
  \label{eq:also-important3}
\end{equation}
Finally, the features obtained from these two modules are concatenated:
\begin{equation}
  F_{SEB} = F_{L} + F_{G}.
  \label{eq:also-important4}
\end{equation}

\subsection{Texture Estimation Branch}
Due to the spatial variability of image deformation and the discontinuity of SCIs, we propose the texture estimation branch (TEB) to predict the features of image deformation effectively. We employ non-uniform B-splines for feature embedding and multiply the estimated local texture by the local Jacobian matrix of the coordinate transformation to predict the features of the distorted image. Specifically, the implicit neural representation of non-uniform B-splines is:
\begin{equation}
  I_{\text{warp}}[y; \theta, \psi] = \sum_{t \in N} w_t f_{\theta}(g_{\psi}(z_t, y - f(x_t), s)),
  \label{eq:eq1}
\end{equation}
where \( z = E_{\phi}(I_{\text{LR}}) \), and \( N \) is a set defined as \( N = \{ N \mid N = [f^{-1}(y) + \left[ \frac{m}{w}, \frac{n}{h} \right]] \mid [m, n] \in [-1, 1] \} \).In this context, \( w_t \) represents the weight of the local set, \( z_t \in \mathbb{R}^D \) is a latent variable for index \( t \), \( x_t \in X \subset \mathbb{R}^2 \) is a coordinate of \( z_t \), and \( s \) is the cell value represented by a superscript factor. The local grid \( \delta_y = y - f(x_t) \) represents the query point \( y = f(x) \in \mathbb{R}^2 \). The function \( g_{\psi}(z_t, \delta_y, s) \) represents the B-spline texture coefficients estimator (BTC), which includes three estimators: the coefficient estimator \( g_c: \mathbb{R}^D \to \mathbb{R}^C \), the structure estimator \( g_k: \mathbb{R}^D \to \mathbb{R}^{2C} \), and the dilation estimator \( g_d: \mathbb{R}^{10} \to \mathbb{R}^C \). The encoding function \( g_{\psi}: (\mathbb{R}^D, \mathbb{R}^2, \mathbb{R}^{10}) \to \mathbb{R}^C \) is defined as :

\begin{equation}
  g_{\psi}(z_t, \delta_y, s) = c_t \odot \text{vec}\left[\beta_n((\delta_{y}-k_{t})\odot d)\right],
  \label{eq:8}
\end{equation}
where \( c_t = g_c(z_t) \), \( k_t = g_k(z_t) \), and \( d = g_d(s) \). To enable the model to represent distorted images, we linearize the given coordinate transformation into an affine transformation:  \( \delta_y = J_f(x_j) \delta_x \). Here,\( J_f(x_j) \in \mathbb{R}^{2 \times 2} \) represents the Jacobian matrix of the coordinate transformation \( f \) at \( x_j \).

Our model can then extract B-spline information from distorted images by utilizing the local grid in the input coordinate space \( \delta_x \) rather than in \( \delta_y \). In arbitrary-scale super-resolution (SR) tasks, the pixels in the upsampled image are square and spatially invariant. However, when the image undergoes warping, the pixels in the resampling image can have arbitrary shapes and spatial transformations. To effectively represent and compute pixel shapes, we represent the pixel shape \( s(y) \in \mathbb{R}^{12} \) (where \( \mathbb{R}^{4} \) denotes pixel direction and \( \mathbb{R}^{4} \ \) denotes pixel curvature) with the gradient of the coordinate transformation at point \( y \) as follows:
\begin{equation}
  s(y) = [J_f^{-1}(y), H_f^{-1}(y)],
  \label{eq:also-important5}
\end{equation}
where \( J_f^{-1}(y) \in \mathbb{R}^{2 \times 2} \) and \( H_f^{-1}(y) \in \mathbb{R}^{2 \times 2 \times 2} \) represent the numerical Jacobian matrix in the pixel direction and the numerical Hessian tensor in the pixel curvature, which are used to describe edge positions and pixel shapes, respectively. 

 Therefore, we redefine the implicit neural representation of non-uniform B-splines in Eq.~\ref{eq:8} as follows:
 \begin{equation}
  F_{TEB} = c_t \odot \text{vec}\left[\beta_n(( J_f(x_j)\delta_{x}-k_{t})\odot g_d(s(y)))\right].
  \label{eq:also-important6}
\end{equation}

\subsection{Structure-Texture Fusion Module}
We design a structure-texture fusion module that integrates the most relevant texture features into structural features to enhance the quality and consistency of texture synthesis. As illustrated in Fig.~\ref{fig2}, we use nearest-neighbor interpolated features sampled from $F_{SEB}$ as queries ($Q$), and utilize $F_{TEB}$ as keys ($K$) and values ($V$). To retrieve texture features most relevant to pixel features $F_{TEB}$, we compute the similarity matrix $R$ between queries $Q$ and keys $K$, where each element $r_{i,j}$ is calculated according to the following formula:
\begin{equation}
  r_{i,j} = \frac{q_i \cdot k_j}{\sqrt{q_i \cdot q_i} \sqrt{k_j \cdot k_j}}.
  \label{eq:also-important7}
\end{equation}
Based on the similarity matrix $R$, we obtain a position index matrix $T$ that identifies the positions of the texture features $k_j$ most similar to each query $q_i$. For each query $q_i$, we find the position with the highest relevance, as represented by the following formula:
\begin{equation}
  t_i = \text{arg} max_j (r_{i,j}),
  \label{eq:also-important8}
\end{equation}
where $t_i$ denotes the position index of the texture feature most similar to the query $q_i$, with values ranging from $1$ to $\frac{h}{2} \times \frac{w}{2}$. Once the most relevant position $t_i$ for each query is determined, we can use this index to extract the corresponding data from the texture branch feature map $V$, obtaining the retrieved texture features $C$ as follows: $c_i = v_{t_i}$.where $c_i$ is an element of the retrieved texture features $C$, and $v_{t_i}$ represents the value at the $t_i$-th position in $V$.
To fuse the retrieved texture features with the structural features $F_{SEB}$, we first concatenate $C$ with $F_{SEB}$ and process them through a feed-forward network to obtain the aggregated features $A$. Finally, we compute the soft attention map $S$, where the elements $s_i$ in $S$ represent the confidence of each element $a_i$ in the retrieved texture features $A$, calculated as follows:
\begin{equation}
  s_i = max_j (r_{i,j}).
  \label{eq:also-important9}
\end{equation}

The final structure-texture fusion features $F_{STF}$ can be obtained using the following formula:
\begin{equation}
  F_{STF} = F_{SEB} + (A \otimes S).
  \label{eq:also-important10}
\end{equation}

\subsection{The Output of STEN Model}
As illustrated in Fig.~\ref{fig2}, two decoders, each implemented by an MLP, are individually used to predict the corresponding outputs based on the fused features and texture features mentioned above. Specifically, we input the texture features \( F_{TEB} \) into the MLP \( f_\theta \), allowing them to be directly decoded into RGB values. Simultaneously, the structure-texture fusion features \( F_{STF} \) are also processed through the MLP \( f_\theta \) to produce additional RGB values. Finally, we sum the RGB values obtained from both decoding processes to generate the final output image \( I_{P_{red}} \). The relationship can be expressed mathematically as follows:
\begin{equation}
  I_{P_{red}} = f_\theta(F_{STF}) + f_\theta(F_{TEB}).
  \label{eq:also-important11}
\end{equation}

\subsection{Transformation Self-estimation Module}
The proposed method requires two inputs: the input warped image \( I \) and the transformation matrix. In practical applications, the transformation matrix is often unknown and needs to be estimated. 
Motivated by blur kernel estimation for blind super-resolution \cite{cornillere2019blind}, we first construct and train a convolutional neural network to predict the transformation error for the estimated transformation matrix based on the warped input image, then we update the estimation of the transformation matrix based on the gradient descent algorithm. This optimization procedure is executed iteratively till the termination condition is satisfied. 
\begin{algorithm}
\SetAlgoLined 
    \caption{Transformation Self-estimation}
    $I_w$: The input warped image\;
    $m_1, m_2, \ldots, m_n$: The initialized transformations\;
    $E$: The well-trained transformation error estimator\;
    $n, T$: The number of samples and iterations, respectively\; 
    $E^*_{\text{min}} \gets \infty$\;
    
    \For{$i \in 1..n$}{
        $I^* \gets \text{STEN}(I_w, m_i)$\;
        $E^* \gets E(I^*)$\;  
        \If{$E^* < E^*_{\text{min}}$}{
            $E^*_{\text{min}} \gets E^*$\;
            $m \gets m_i$\;
        }
    }
    
    \For{$j \in 1..T$}{
        $I^* \gets \text{STEN}(I_w, m)$\;
        $e \gets E(I^*)$\;   
        $loss \gets e + \alpha \cdot |m|$\;
        $loss.\text{backward}(m)$\;  
    }    

     \textbf{Output:} {$m$}\;
\label{alg1}
\end{algorithm}

\begin{table*}[t]
\renewcommand{\arraystretch}{1.05} 
\begin{center}
\vspace{-0.5em}
\caption{Quantitative comparison of arbitrary scale super-resolution methods on SCI1K, SCID, and SIQAD datasets within in-scale (PSNR (dB)). The best results are highlighted in bold, while the second-best results are indicated with an underline.}
 \setlength{\tabcolsep}{10pt}{
  \begin{tabular}{ll llc llcc}
    \begin{tabular}{ll llc llcc}
\toprule
\multicolumn{2}{c}{ Train set: SCI1K (n=800)} & \multicolumn{3}{c}{ In-training-scale} & \multicolumn{4}{c}{ Out-of-training-scale} \\
\cmidrule(lr){1-2} \cmidrule(lr){3-5} \cmidrule(lr){6-9}
\multicolumn{1}{c}{Test set} &  Method & \multicolumn{1}{c}{ $\times2$} & \multicolumn{1}{c}{ $\times3$}& \multicolumn{1}{c}{ $\times4$} & \multicolumn{1}{c}{ $\times5$} & \multicolumn{1}{c}{ $\times6$} & \multicolumn{1}{c}{ $\times7$} & \multicolumn{1}{c}{ $\times8$}  \\
\midrule
\midrule
\multirow{8}{*}{\parbox{2cm}{\centering SCI1K \\ (n = 200)}} 
& \raggedright Bicubic & 28.81 & 25.15 & 23.18 & 22.02 & 21.23 & 20.72 & 20.26  \\
& \raggedright RDN \cite{zhang2018residual}  & 38.45 & 33.59 & 29.81 & \multicolumn{1}{c}{-} & \multicolumn{1}{c}{-} & \multicolumn{1}{c}{-} & \multicolumn{1}{c}{-} \\
& \raggedright MetaSR \cite{hu2019meta} & 38.57 & 33.67 & 30.12 & 27.52 & 26.13 & 23.91 & 23.19  \\
& \raggedright LIIF  \cite{chen2021learning}  & 38.65 & 33.97 & 30.55 & 27.77 & 26.07 & 23.99 & 23.24  \\
& \raggedright LTE \cite{lee2022local} & 39.14 & 34.50 & 30.93 & 28.22 & 26.19 & 24.28 & 23.17 \\
& \raggedright BTC\cite{pak2023b}  & \underline {39.17} & \underline{34.58} & \underline{31.10} & \underline{28.33} &\underline{26.31} & \textbf{24.47} &\underline{ 23.38}  \\
& \raggedright STEN(Ours) &  \textbf{39.22 }&  \textbf{34.87} &  \textbf{31.33} & \textbf{28.34} &  \textbf{26.38} &\underline{24.37} &\textbf{23.40} \\
\midrule
\multirow{ 8}{*}{\parbox{2cm}{\centering SCID \\ (n = 40)}} 
&  Bicubic & 25.22 & 22.78 & 21.60 & 20.90 & 20.42 & 20.04 & 19.77  \\
&  RDN \cite{zhang2018residual}  & 34.00 & 28.34 & 25.74 & \multicolumn{1}{c}{-} & \multicolumn{1}{c}{-} & \multicolumn{1}{c}{-}& \multicolumn{1}{c}{-} \\
&  MetaSR \cite{hu2019meta} & 33.84 & 29.08 & 25.76 & 23.62 & 22.38 & 21.59 & 21.07 \\
&  LIIF  \cite{chen2021learning}  & 34.24 & 29.10 & 25.89 & 23.77 & 22.53 & 21.73 & 21.21 \\
&  LTE \cite{lee2022local} &\underline{34.49} & \underline{29.60} & \underline{26.34} & 24.06 & \underline{22.67} & 21.81 & 21.28  \\
&  BTC\cite{pak2023b}   & 34.48 & 29.56 & 26.30 & \underline{24.09} & \textbf{22.69} & \underline{21.84} & \textbf{21.29} \\
&  STEN(Ours)   & \textbf{34.59 } & \textbf{29.81 } & \textbf{26.60} & \textbf{24.15} & 22.65 & \textbf{21.85}&\underline{21.28} \\
\midrule
\multirow{8}{*}{\parbox{2cm}{\centering SCIAQ \\ (n = 22)}} \textbf
&  Bicubic & 22.89 & 20.66 & 19.70 & 19.18 & 18.79 & 18.46 & 18.20 \\
& RDN \cite{zhang2018residual}  & 33.53 & 26.89 & 23.38 & \multicolumn{1}{c}{-} & \multicolumn{1}{c}{-} & \multicolumn{1}{c}{-} & \multicolumn{1}{c}{-} \\
&  MetaSR \cite{hu2019meta} & 34.12 & 28.40 & 23.55 & 21.18 & 20.18 & 19.63 & 19.25 \\
&  LIIF  \cite{chen2021learning}  & 34.31 & 28.27 & 23.44 & 21.16 & 20.25 & 19.70 & 19.36\\
&  LTE \cite{lee2022local} & \underline{35.07} & 29.33 & 24.21 & 21.52 & 20.39 & 19.78 & 19.43  \\
&  BTC \cite{pak2023b}  & 34.91 & \underline{29.36} & \underline{24.25} & \underline{21.57} &\textbf{20.43} & \underline{19.82} & \underline{19.45}  \\
&  STEN(Ours)   & \textbf{35.20} & \textbf{29.71 } & \textbf{24.60} & \textbf{21.63} & \underline{20.40} & \textbf{19.85}&\textbf{19.48} \\
\bottomrule
\end{tabular}
\end{tabular}}
\end{center}

\label{tab1}
\vspace{-0.5em}
\end{table*}

During the training of the transformation error estimation CNN, we treat the difference between the image generated by the estimated transformation and the warped input image as the loss function. This approach ensures that the input image can be restored to its original state after undergoing cycle-consistent translations \cite{zhu2017unpaired}, maintaining feature consistency between the input and output.

Once the transformation error estimator (error estimation CNN) is well-trained, we can estimate the unknown transformation in real-world scenarios. Concretely, given a warped image $I$, the transformation error estimator can obtain the transformation difference. Based on the difference, we can update the estimated transformation matrix iteratively and finally get a good approximation of the genuine transformation. Assuming $M$ is the estimated transformation matrix, the transformation estimation optimization can be described as:
\begin{equation}
  M^* = \arg\min_{M} \left( E(S, M, I) + \alpha \left| M \right| \right) \ ,
  \label{eq:also-important13}
\end{equation}
where  \(E \) represents the transformation error estmatior, \( S \) is our proposed unwarping model (STEN), \( \left| M \right| \) denotes the size of the matrix \( M \), and \( \alpha \geq 0 \) is a tunable parameter. The description of the transformation self-estimation algorithm is detailed in Algorithm~\ref {alg1}.

\section{Experiments}
\subsection{Implementation Details}
\textbf{Dataset }
We use the existing SCI1K, SCID, and SCIAQ dataset \cite{yang2021implicit} to construct our warped dataset - SCI1KW, SCIDW, and SCIAQW  to train and test our STEN model. Specifically, we first randomly generate a warping matrix, which includes transformations such as random scaling, cropping, rotation, and projection. Then, we distort each image in SCI1K with the warping matrix to obtain a distorted version and finally form the SCI1KW datasets. During training processing, we crop the maximum effective region of the warped result as the input image according to the transformation effect. In the testing phase, we centre-crop the \(384\times384 \) size patch from each test image in the test set and assign a transformation on the patch. This method allows us to effectively evaluate the model's performance under different warping conditions.
\textbf{Settings} 
For optimization, we employ L1 loss \cite{lim2017enhanced} and the Adam optimizer \cite{kingma2014adam} with $\beta_1 = 0.9$ and $\beta_2 = 0.999$. Our STEN model is trained for 1000 epochs with a batch size of 16, starting with a learning rate of $1 \times 10^{-4}$, halving it every 200 epochs. In the transformation self-estimation algorithm, we set the adjustable parameter $\ alpha=0.05$. 
All the compared methods used in this work adopt the same settings as our STEN model.

\begin{figure*}[!ht]
    \centering
    \rotatebox{90}{\scriptsize \textbf{ ~~~~~~~~~~~x2}}
    \begin{minipage}[t]{0.1755\textwidth} 
        \centering        \includegraphics[width=\textwidth,height=1.75cm]{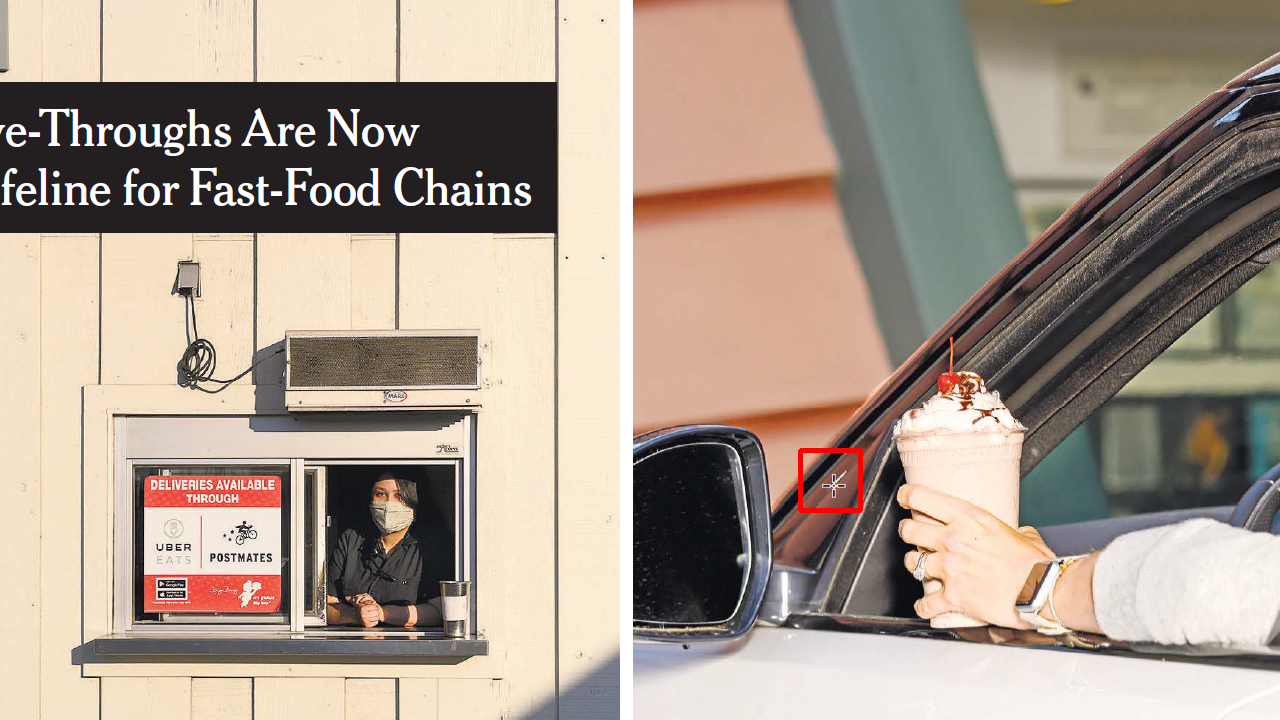}
    \end{minipage}    
    \begin{minipage}[t]{0.1\textwidth} 
        \centering
        \includegraphics[width=\textwidth]{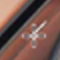}
    \end{minipage}
    \begin{minipage}[t]{0.1\textwidth}
        \centering
        \includegraphics[width=\textwidth]{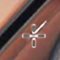}
    \end{minipage}
    \begin{minipage}[t]{0.1\textwidth}
        \centering
        \includegraphics[width=\textwidth]{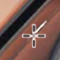}
    \end{minipage}
    \begin{minipage}[t]{0.1\textwidth}
        \centering
        \includegraphics[width=\textwidth]{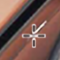}
    \end{minipage}
    \begin{minipage}[t]{0.1\textwidth}
        \centering
        \includegraphics[width=\textwidth]{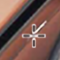}
    \end{minipage}
    \begin{minipage}[t]{0.1\textwidth}
        \centering
        \includegraphics[width=\textwidth]{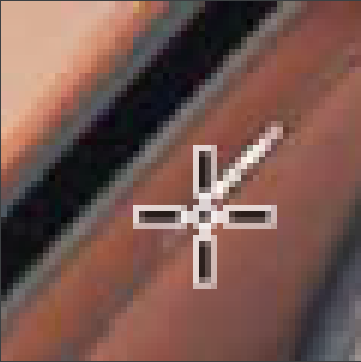}
    \end{minipage}
    \begin{minipage}[t]{0.1\textwidth}
        \centering
        \includegraphics[width=\textwidth]{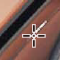}
    \end{minipage}
    \vspace{0.5mm}  

    \rotatebox{90}{\scriptsize \textbf{~~~~~~~~~~~x3.5}}
    \begin{minipage}[t]{0.1755\textwidth} 
        \centering
        \includegraphics[width=\textwidth,height=1.75cm]{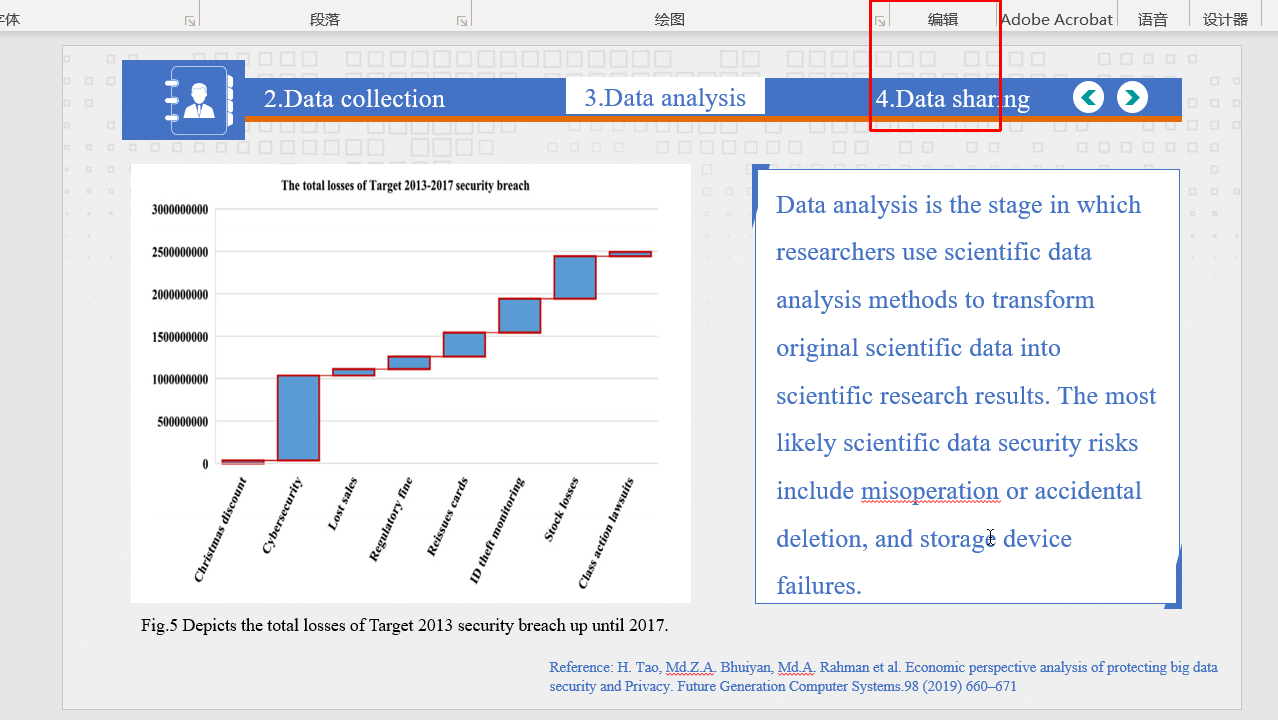}
        \vspace{\baselineskip}
       \centerline {HR Image} 
    \end{minipage}    
    \begin{minipage}[t]{0.10\textwidth} 
        \centering
        \includegraphics[width=\textwidth]{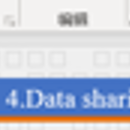}
       \vspace{\baselineskip}
        \centerline{Bicubic} 
    \end{minipage}
    \begin{minipage}[t]{0.10\textwidth}
        \centering
        \includegraphics[width=\textwidth]{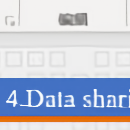}
        \vspace{\baselineskip}
        \centerline{MetaSR}
    \end{minipage}
    \begin{minipage}[t]{0.10\textwidth}
        \centering
        \includegraphics[width=\textwidth]{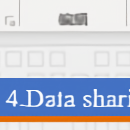}
        \vspace{\baselineskip}
        \centerline{LIIF}
    \end{minipage}
    \begin{minipage}[t]{0.10\textwidth}
        \centering
        \includegraphics[width=\textwidth]{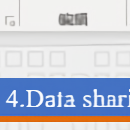}
        \vspace{\baselineskip}
        \centerline{LTE}
    \end{minipage}
    \begin{minipage}[t]{0.10\textwidth}
        \centering
        \includegraphics[width=\textwidth]{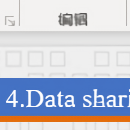}
        \vspace{\baselineskip}
        \centerline{BTC}
    \end{minipage}
    \begin{minipage}[t]{0.10\textwidth}
        \centering
        \includegraphics[width=\textwidth]{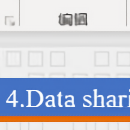}
        \vspace{\baselineskip}
        \centerline{STEN(Ours)} 
    \end{minipage}
    \begin{minipage}[t]{0.10\textwidth}
        \centering
        \includegraphics[width=\textwidth]{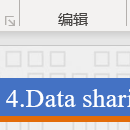}
        \vspace{\baselineskip}        
        \centerline{GT}   
    \end{minipage}   
     \vspace{-1em}
    \caption{Qualitatively compare to other arbitrary scale super-resolution (SR) methods, \emph{i.e.,} MetaSR~\cite{hu2019meta}, LIIF\cite{chen2021learning}, LTE\cite{lee2022local}, and BTC\cite{pak2023b}, at scales $\times2$ and $\times3.5$ within in scale.
    }
    \label{fig3}
\end{figure*}
\begin{figure*}[!ht]
    \centering
       \rotatebox{90}{\scriptsize \textbf{ ~~~~~~~~~~~x5}}
    \begin{minipage}[t]{0.1755\textwidth} 
        \centering        \includegraphics[width=\textwidth,height=1.75cm]{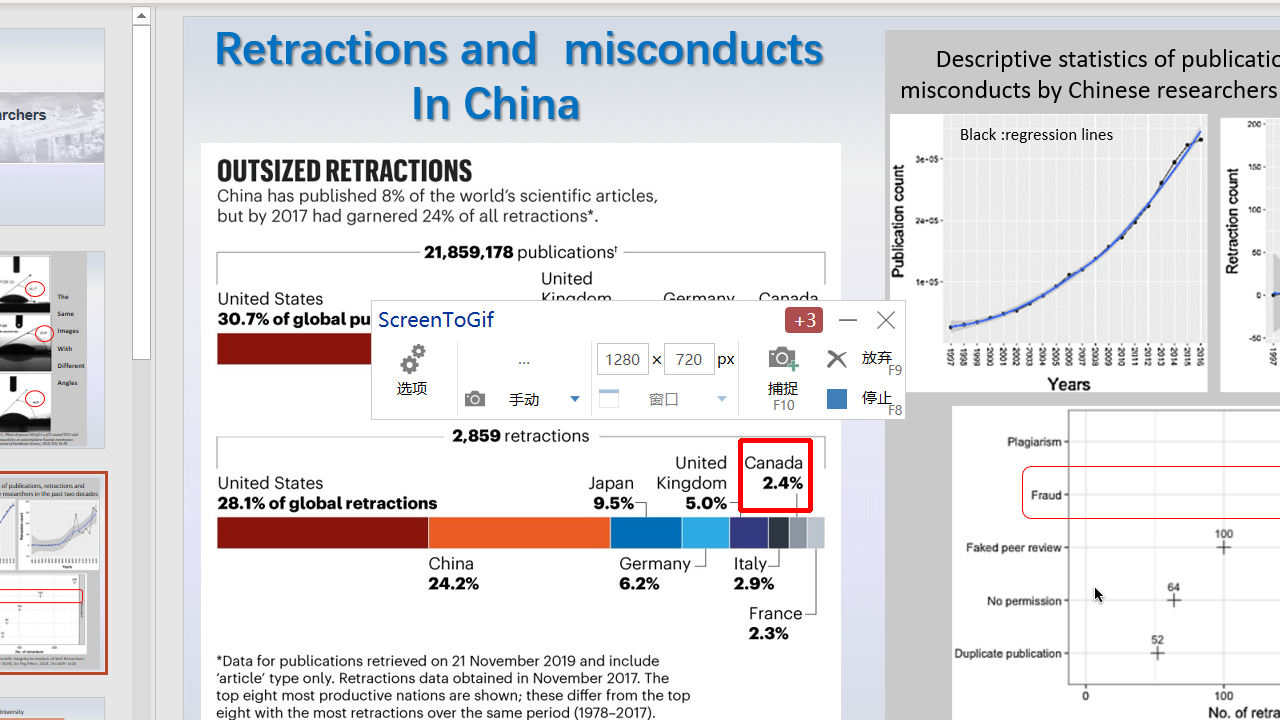}
    \end{minipage}    
    \begin{minipage}[t]{0.10\textwidth} 
        \centering
        \includegraphics[width=\textwidth]{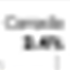}
    \end{minipage}
    \begin{minipage}[t]{0.10\textwidth}
        \centering
        \includegraphics[width=\textwidth]{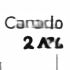}
    \end{minipage}
    \begin{minipage}[t]{0.10\textwidth}
        \centering
        \includegraphics[width=\textwidth]{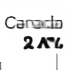}
    \end{minipage}
    \begin{minipage}[t]{0.10\textwidth}
        \centering
        \includegraphics[width=\textwidth]{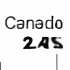}
    \end{minipage}
    \begin{minipage}[t]{0.10\textwidth}
        \centering
        \includegraphics[width=\textwidth]{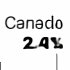}
    \end{minipage}
    \begin{minipage}[t]{0.10\textwidth}
        \centering
        \includegraphics[width=\textwidth]{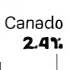}
    \end{minipage}
    \begin{minipage}[t]{0.10\textwidth}
        \centering
        \includegraphics[width=\textwidth]{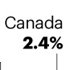}
    \end{minipage}   
    \vspace{0.5mm}

    \rotatebox{90}{\scriptsize \textbf{ ~~~~~~~~~~~x6.4}}
    \begin{minipage}[t]{0.1755\textwidth}
        \centering
        \includegraphics[width=\textwidth,height=1.75cm]{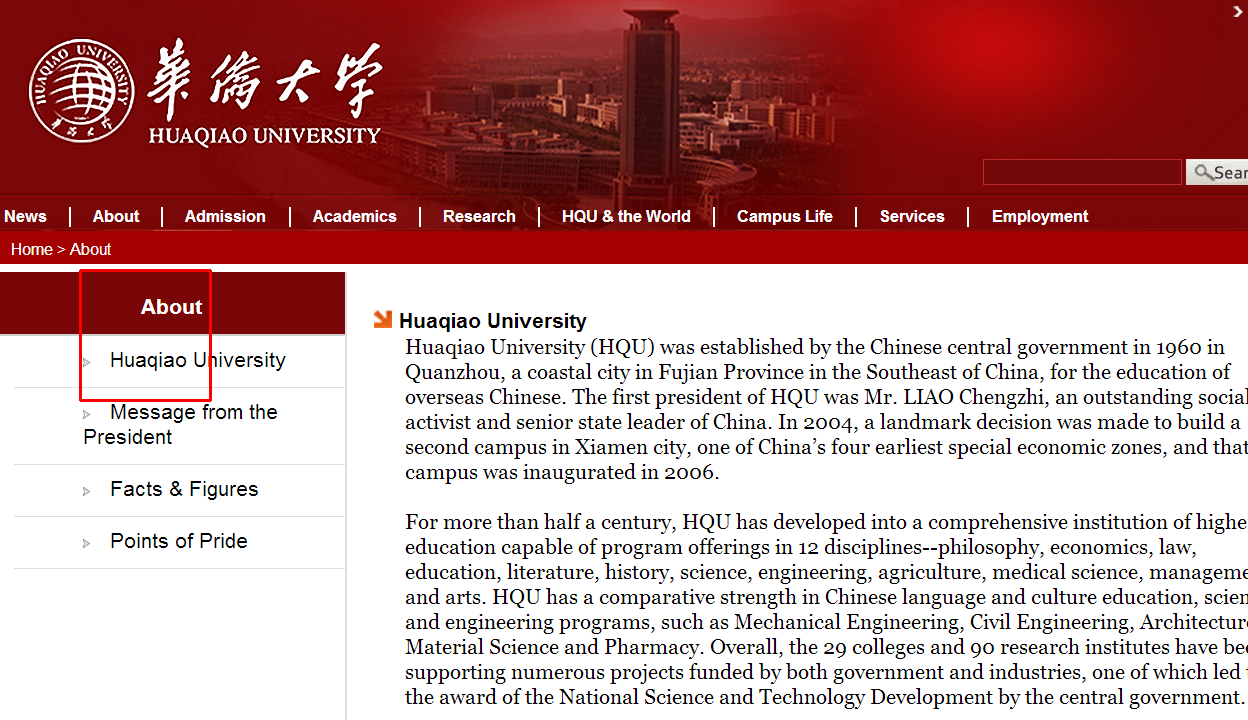}
        \vspace{\baselineskip}
       \centerline {HR Image} 
    \end{minipage}    
    \begin{minipage}[t]{0.10\textwidth} 
        \centering 
        \includegraphics[width=\textwidth]{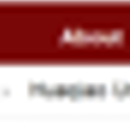}
        \vspace{\baselineskip}
        \centerline{Bicubic}
    \end{minipage}
    \begin{minipage}[t]{0.10\textwidth}
        \centering
        \includegraphics[width=\textwidth]{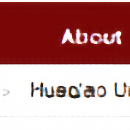}
        \vspace{\baselineskip}
        \centerline{MetaSR}
    \end{minipage}
    \begin{minipage}[t]{0.10\textwidth}
        \centering
        \includegraphics[width=\textwidth]{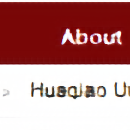}
        \vspace{\baselineskip}
        \centerline{LIIF}
    \end{minipage}
    \begin{minipage}[t]{0.10\textwidth}
        \centering
        \includegraphics[width=\textwidth]{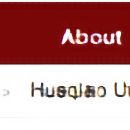}
        \vspace{\baselineskip}
        \centerline{LTE}
    \end{minipage}
    \begin{minipage}[t]{0.10\textwidth}
        \centering
        \includegraphics[width=\textwidth]{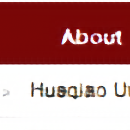}
        \vspace{\baselineskip}
        \centerline{BTC}
    \end{minipage}
    \begin{minipage}[t]{0.10\textwidth}
        \centering
        \includegraphics[width=\textwidth]{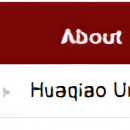}
        \vspace{\baselineskip}
        \centerline{STEN(Ours)}
    \end{minipage}
    \begin{minipage}[t]{0.10\textwidth}
        \centering
        \includegraphics[width=\textwidth]{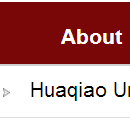}
        \vspace{\baselineskip}        
        \centerline{GT}    
    \end{minipage}   
    \vspace{-1em}
    \caption{Qualitatively compare to other arbitrary scale super-resolution (SR) methods, \emph{i.e.,} MetaSR~\cite{hu2019meta}, LIIF\cite{chen2021learning}, LTE\cite{lee2022local}, and BTC\cite{pak2023b}, at scales $\times2$ and $\times3.5$ within in scale. at scales $\times5$ and $\times6.4$ out of scale. 
    }
    \label{fig4}
    \vspace{-0.5em}
\end{figure*}

\subsection{Comparisons for Arbitrary Scale Super-Resolution}
We compare our proposed STEN model with BTC\cite{pak2023b}, LTE\cite{lee2022local}, LIIF\cite{chen2021learning}, and MetaSR\cite{hu2019meta} for arbitrary-scale image super-resolution in both in-scale and out-of-scale scenarios. To ensure a fair comparison, all methods use RDN\cite{zhang2018residual} as the encoder for feature extraction and are trained on the SCI1K dataset. As shown in Table~\ref{tab1}, our method outperforms existing arbitrary-scale super-resolution techniques across most scale factors and datasets.

Additionally, Fig.~\ref{fig3} and Fig.~\ref{fig4} show a qualitative comparison between our method and the others, demonstrating reconstruction performance at different scales. Whether in-scale or out-of-scale, our method consistently surpasses others in recovering text and graphical details, highlighting its clear advantages in detail preservation and visual quality. 

\begin{table}[t]
\renewcommand{\arraystretch}{1.1}
 \begin{center}
 \vspace{-0.5em}
\caption{Quantitative comparison of homography transform methods on SCI1KW, SCIDW, and SIQADW datasets within in-scale (isc) and out-of-scale (osc) (PSNR (dB)). The best results are highlighted in bold, while the second-best results are indicated with an underline.}
 \setlength{\tabcolsep}{1.4pt}{
\begin{tabular}{l|cc|cc|cc}
\toprule
\multirow{2}{*}{Method} & \multicolumn{2}{c|}{SCI1KW} & \multicolumn{2}{c|}{SCIDW}  & \multicolumn{2}{c}{SIQADW} \\
 & isc & osc & isc & osc & isc & osc \\
\hline
Bicubic & 24.93  & 23.04  & 22.43  & 21.02  & 20.36  & 19.21  \\
\hline
RDN\cite{zhang2018residual} & 33.81  & 25.08  & 28.54  & 22.39  & 27.24  & 21.04  \\
SRWarp-RDN\cite{son2021srwarp} & 35.93  & 27.65  & 31.65  & 24.38  & 31.24  & 21.89  \\
LTEW-RDN\cite{lee2022learning} & \underline{36.18} & \underline{28.55} & \underline{32.17} & \underline{25.91} & \underline{31.67} & \underline{22.31} \\
STEN-RDN(est) & 25.84 & 22.78 & 22.94 & 20.68 & 20.43& 18.62 \\
STEN-RDN(Ours) & \textbf{36.97} & \textbf{29.29} & \textbf{32.49} & \textbf{26.17} & \textbf{32.15 } &\textbf{22.59 } \\
\bottomrule
\end{tabular}
}
\label{tab2}
\end{center}
\vspace{-2.5em}
\end{table}

\begin{figure*}[!ht]
    \centering
    \begin{minipage}[t]{0.20\textwidth} 
        \centering        \includegraphics[width=\textwidth,height=2.15cm]{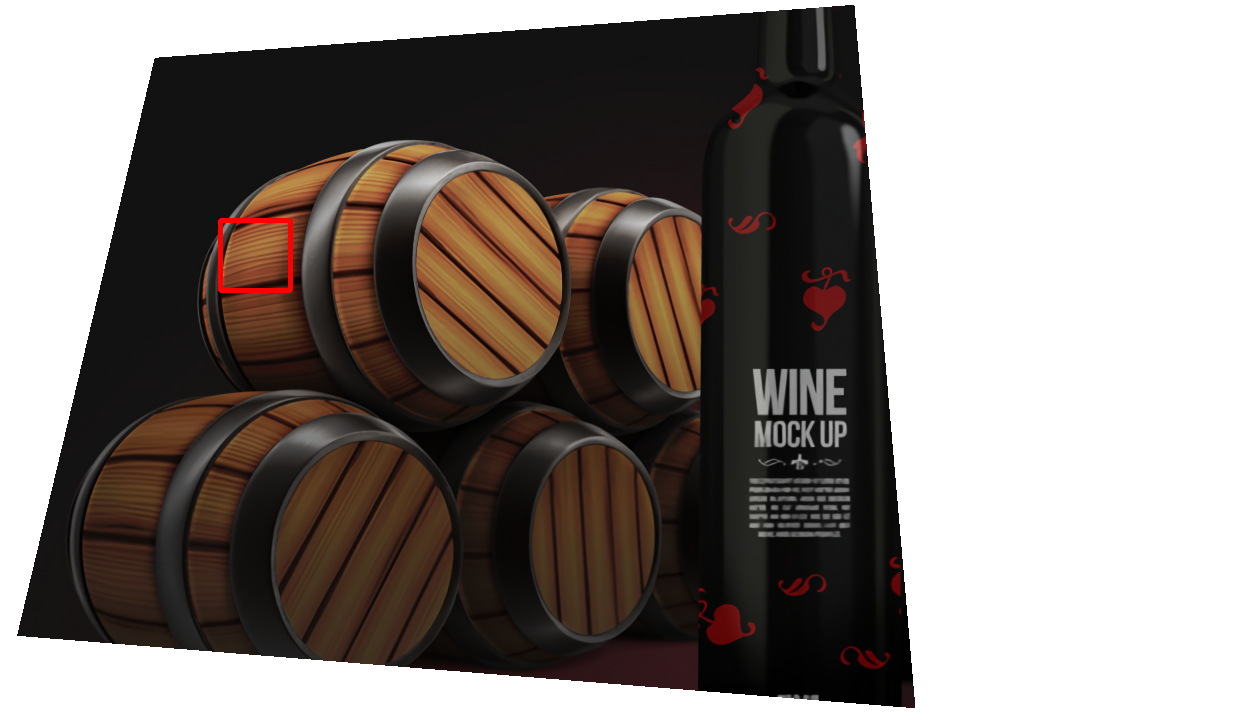}
    \end{minipage}    
    \begin{minipage}[t]{0.12\textwidth}
        \centering
        \includegraphics[width=\textwidth]{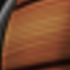}
    \end{minipage}
    \begin{minipage}[t]{0.12\textwidth}
        \centering
        \includegraphics[width=\textwidth]{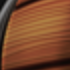}
    \end{minipage}
    \begin{minipage}[t]{0.12\textwidth}
        \centering
        \includegraphics[width=\textwidth]{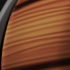}
    \end{minipage}
    \begin{minipage}[t]{0.12\textwidth}
        \centering
        \includegraphics[width=\textwidth]{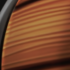}
    \end{minipage}
    \begin{minipage}[t]{0.12\textwidth}
        \centering
        \includegraphics[width=\textwidth]{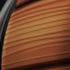}
    \end{minipage}
    \begin{minipage}[t]{0.12\textwidth}
        \centering
        \includegraphics[width=\textwidth]{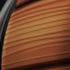}
    \end{minipage}   
    \vspace{0.5mm}
  
    \begin{minipage}[t]{0.2\textwidth} 
        \centering
        \includegraphics[width=\textwidth,height=2.15cm]{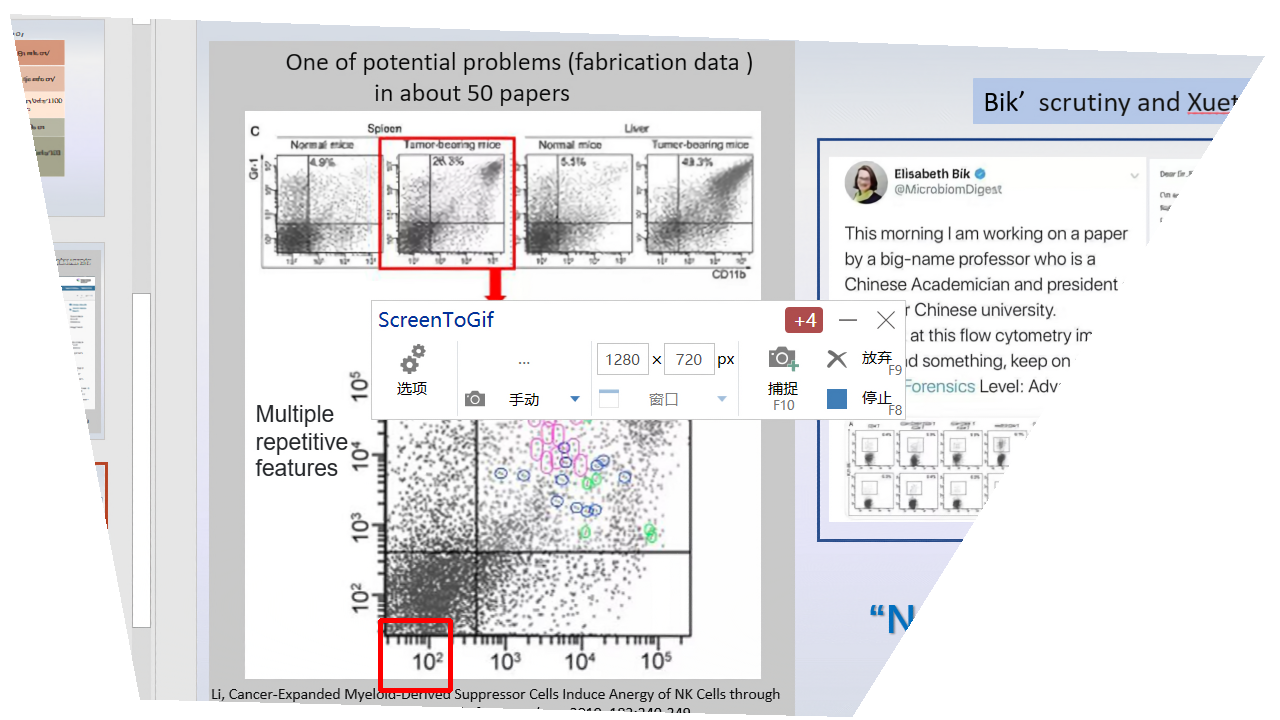}
        \vspace{\baselineskip}
        \centerline{HR Image} 
    \end{minipage}    
    \begin{minipage}[t]{0.12\textwidth}
        \centering 
        \includegraphics[width=\textwidth]{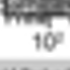}
        \vspace{\baselineskip}
        \centerline{Bicubic}
    \end{minipage}
    \begin{minipage}[t]{0.12\textwidth}
        \centering
        \includegraphics[width=\textwidth]{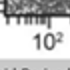}
        \vspace{\baselineskip}
        \centerline{RDN}
    \end{minipage}
    \begin{minipage}[t]{0.12\textwidth}
        \centering
        \includegraphics[width=\textwidth]{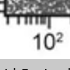}
        \vspace{\baselineskip}
        \centerline{SRWarp}
    \end{minipage}
    \begin{minipage}[t]{0.12\textwidth}
        \centering
        \includegraphics[width=\textwidth]{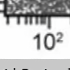}
        \vspace{\baselineskip}
        \centerline{LTEW}
    \end{minipage}
    \begin{minipage}[t]{0.12\textwidth}
        \centering
        \includegraphics[width=\textwidth]{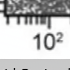}
        \vspace{\baselineskip}
        \centerline{STEN (Ours)} 
    \end{minipage}
    \begin{minipage}[t]{0.12\textwidth}
        \centering
        \includegraphics[width=\textwidth]{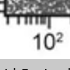}
        \vspace{\baselineskip}
        \centerline{GT} 
    \end{minipage}
    \vspace{-1em}
    \caption{Qualitative comparison to other homography transform  methods, \emph{i.e.,} RDN\cite{zhang2018residual}, SRWarp\cite{son2021srwarp}, LTEW \cite{lee2022learning} within in-scale.}
    \label{fig5}
\end{figure*}

\begin{figure*}[!ht]
    \centering
    \begin{minipage}[t]{0.2\textwidth} 
        \centering        \includegraphics[width=\textwidth,height=2.15cm]{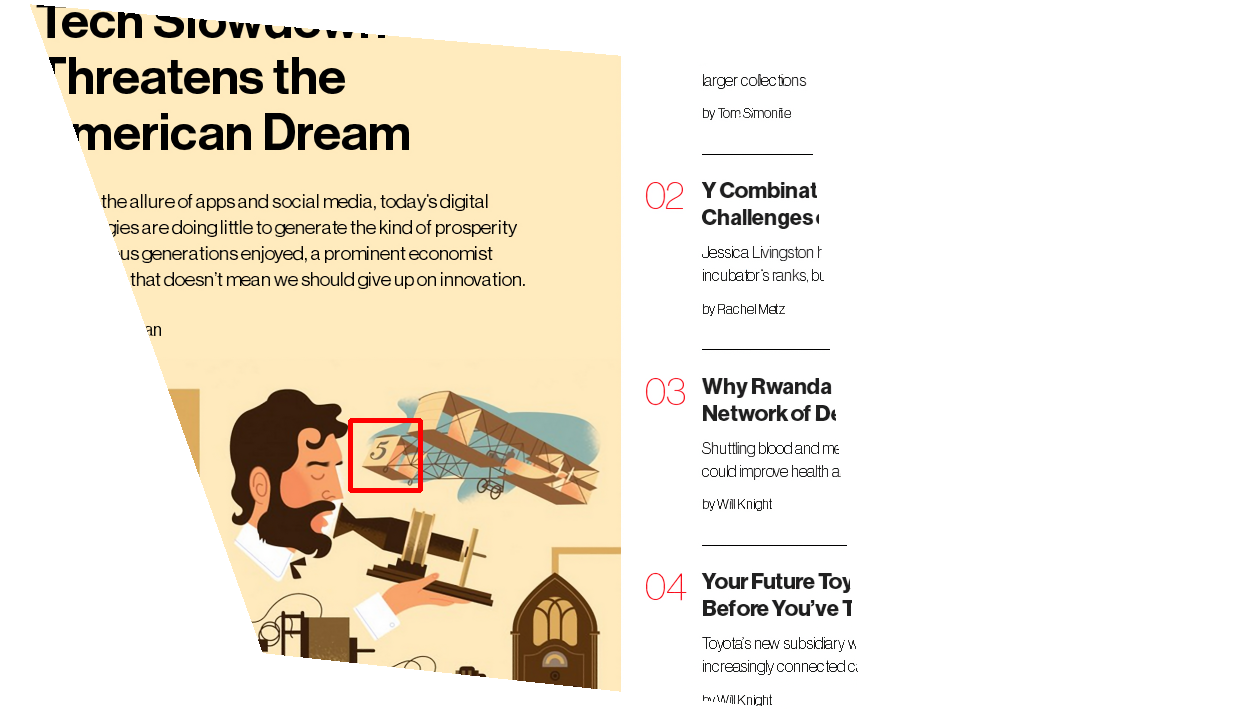}
    \end{minipage}    
    \begin{minipage}[t]{0.12\textwidth}
        \centering
        \includegraphics[width=\textwidth]{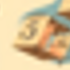}
    \end{minipage}
    \begin{minipage}[t]{0.12\textwidth}
        \centering
        \includegraphics[width=\textwidth]{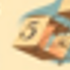}
    \end{minipage}
    \begin{minipage}[t]{0.12\textwidth}
        \centering
        \includegraphics[width=\textwidth]{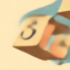}
    \end{minipage}
    \begin{minipage}[t]{0.12\textwidth}
        \centering
        \includegraphics[width=\textwidth]{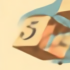}
    \end{minipage}
    \begin{minipage}[t]{0.12\textwidth}
        \centering
        \includegraphics[width=\textwidth]{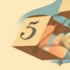}
    \end{minipage}
    \begin{minipage}[t]{0.12\textwidth}
        \centering
        \includegraphics[width=\textwidth]{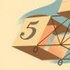}
    \end{minipage}   
    \vspace{0.5mm}
  
    \begin{minipage}[t]{0.2\textwidth} 
        \centering
        \includegraphics[width=\textwidth,height=2.15cm]{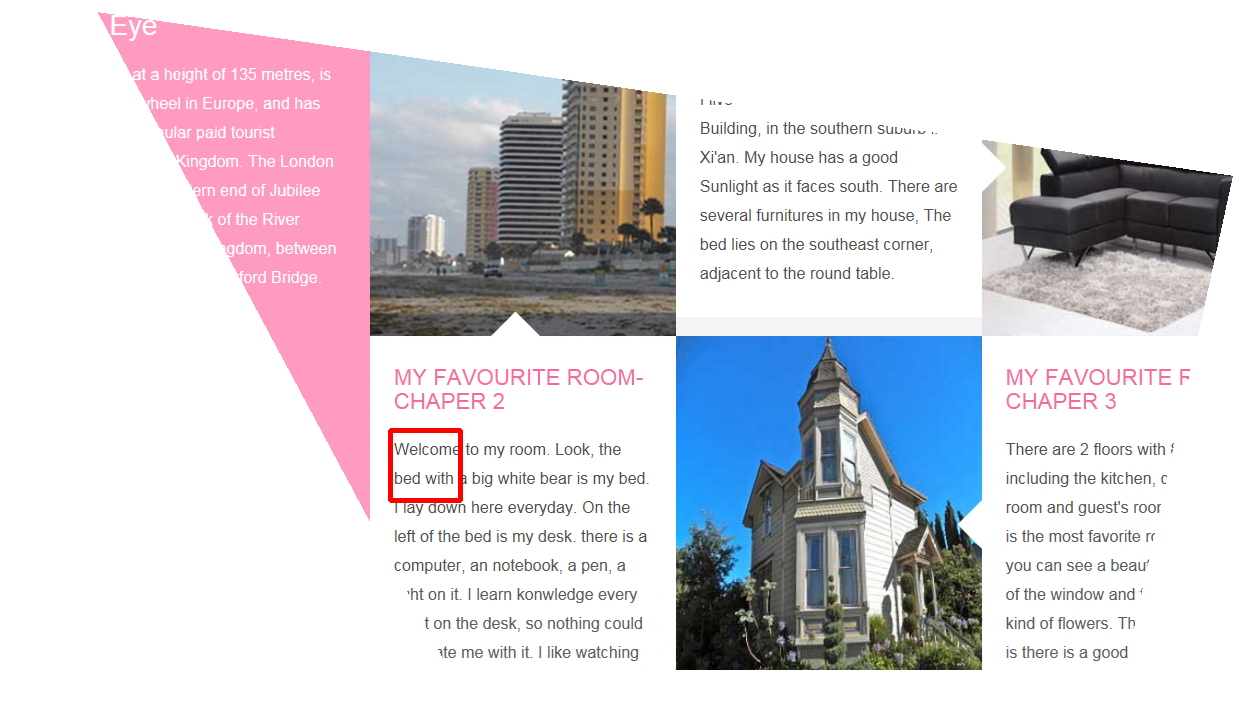}
        \vspace{-1.8\baselineskip}
        \centerline{HR Image}
    \end{minipage}    
    \begin{minipage}[t]{0.12\textwidth} 
        \centering 
        \includegraphics[width=\textwidth]{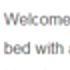}
        \vspace{\baselineskip}
        \centerline{Bicubic} 
    \end{minipage}
    \begin{minipage}[t]{0.12\textwidth}
        \centering
        \includegraphics[width=\textwidth]{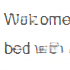}
        \vspace{\baselineskip}
        \centerline{RDN}
    \end{minipage}
    \begin{minipage}[t]{0.12\textwidth}
        \centering
        \includegraphics[width=\textwidth]{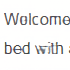}
        \vspace{\baselineskip}
        \centerline{SRWarp}
    \end{minipage}
    \begin{minipage}[t]{0.12\textwidth}
        \centering
        \includegraphics[width=\textwidth]{images/warp/out2/ltew.png}
        \vspace{\baselineskip}
        \centerline{LTEW}
    \end{minipage}
    \begin{minipage}[t]{0.12\textwidth}
        \centering
        \includegraphics[width=\textwidth]{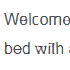}
        \vspace{\baselineskip}
        \centerline{STEN(Ours)}
    \end{minipage}
    \begin{minipage}[t]{0.12\textwidth}
        \centering
        \includegraphics[width=\textwidth]{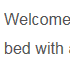}
        \vspace{\baselineskip}
        \centerline{GT} 
    \end{minipage}   
    \vspace{-1em}
    \caption{Qualitative comparison to other homography transform  methods, \emph{i.e.,} RDN\cite{zhang2018residual}, SRWarp\cite{son2021srwarp}, LTEW \cite{lee2022learning} within out-scale.}
    \label{fig6}
    \vspace{-0.5em}
\end{figure*}

\subsection{Comparisons for Homography Warping }
We compare our STEN model with SRWarp\cite{son2021srwarp} and LTEW\cite{lee2022learning} on the generated benchmark datasets (SCI1KW, SCIDW, and SIQADW) with homography warping transformation, including in-distribution and out-of-distribution scenarios. In the in-scale scenario, we consider the scale factors present in the training dataset, while in the out-of-scale scenario, the scale factors were not included in the training dataset. To ensure a fair comparison, all methods use RDN\cite{zhang2018residual} as the encoder for feature extraction and are trained on the SCI1K dataset. Additionally, we include results from bilinear interpolation and the RDN model. For RDN, we upsample the input images and use the WarpPerspective function\cite{bradski2000opencv} with a bicubic kernel to synthesize the warped images. For SRWarp and LTEW, we implement them using their publicly available source code. 

Table~\ref{tab2} presents the average mPSNR results of our method compared to other methods on benchmark datasets under both in-scale and out-of-scale conditions. It can be observed that our method outperforms LTEW\cite{lee2022learning} in both cases. Additionally, we validate the effectiveness of the transformation self-estimation model, where STEN(est) is the result obtained using the estimated transformation matrix \(M_i\). From Table~\ref{tab2}, it is evident that although the performance metrics of STEN(est) are lower, it can still effectively predict the transformation matrix. Furthermore, Fig.~\ref{fig5} and Fig.~\ref{fig6} provide qualitative comparisons of in-scale and out-of-scale methods, showing that our approach generates text and texture details more clearly. In summary, our method outperforms existing homography transformation techniques in terms of both mean Peak Signal-to-Noise Ratio (mPSNR) and visual quality.

\begin{table}[t]
\centering
\caption{For structure estimation branch (SEB) and Structure-Texture Fusion Module (STF), as well as B-spline representation and Fourier representation, quantitative ablation experiments (mPSNR (dB)) are conducted for homography transformations within in-scale (isc) and out-of-scale (osc) scenarios on the SCI1K dataset.}
\setlength{\tabcolsep}{4.5pt}{
\begin{tabular}{ c c c c c | c c }
\hline
\multirow{2}{*}{TEB} & \multirow{2}{*}{SEB} & \multirow{2}{*}{STF}& \multirow{2}{*}{ B-splines}& \multirow{2}{*}{Fourier} & \multicolumn{2}{c}{SCI1KW} \\ 
& & & & & isc & osc \\ \hline
\checkmark & \ding{55} &\ding{55} &\checkmark &\ding{55} & 36.65\ & 29.01\ \\
\checkmark & \ding{55}&\checkmark &\checkmark &\ding{55} & 36.80\ & 29.14 \\
\checkmark & \checkmark &\ding{55} &\checkmark &\ding{55} & 36.79 & 29.10 \\
\checkmark & \checkmark &\checkmark &\ding{55} &\checkmark & 36.83 & 29.13 \\
\checkmark & \checkmark &\checkmark & \checkmark & \ding{55} & \textbf{36.97} & \textbf{29.29} \\\hline
\end{tabular}
}
\label{tab3}
\end{table}
\vspace{-1em}

\begin{table*}
\renewcommand{\arraystretch}{1.05}
\begin{center}
\caption{Quantitative comparison of homography transform methods on natural image benchmarks: DIV2KW, Set5W, Set14W, B100W, and Urban100W (PSNR (dB)). The best results are highlighted in bold, while the second-best results are indicated with an underline.}
\setlength{\tabcolsep}{7.5pt}{
\begin{tabular}{l|cc|cc|cc|cc|cc}
\hline
\multirow{2}{*}{Method} & \multicolumn{2}{c|}{DIV2K} & \multicolumn{2}{c|}{Set5W}  & \multicolumn{2}{c|}{Set14W}& \multicolumn{2}{c|}{B100W} & \multicolumn{2}{c}{Urban100W}\\
 & isc & osc & isc & osc & isc & osc& isc & osc& isc & osc \\
\hline
Bicubic & 27.85  & 25.03  & 35.00  & 28.75  & 28.79  & 24.57& 28.67  & 25.02& 24.84  & 21.89  \\
\hline
RRDB\cite{wang2018esrgan}  & 30.76  & 26.84  & 37.40  & 30.34  & 31.56  & 25.95 & 30.29  & 26.32  & 28.83  & 23.94  \\
SRWarp-RRDB\cite{son2021srwarp} & 31.04  & 26.75  & 37.93  & 29.90  & 32.11  & 25.35 & 30.48  & 26.10  & 29.45  & 24.04  \\
LTEW-RRDB\cite{lee2022learning} &31.10 & 26.92 & \underline{38.20}& 31.07 & 32.15 & 26.02& 30.56  & 26.41 & 29.50  & 24.25 \\
MFR-RRDB\cite{xiao2024towards} & \underline{31.18}  & \underline{27.12}  & \textbf{38.23}  & \textbf{31.19}  & \underline{32.26}  & \underline{26.26} & \underline{30.62}  & \textbf{26.53}  & \textbf{29.68}  & \textbf{24.51}  \\
STEN-RRDB(Ours) & \textbf{31.96} & \textbf{27.75} &38.12 & \underline{31.10}& \textbf{32.30 } &\textbf{26.27} & \textbf{30.88}& \underline{26.45}&\underline{ 29.50}& \underline{24.26}\\
\bottomrule
\end{tabular}}
\end{center}
\label{tab:tabel4}
\end{table*}

\subsection{Ablation Study}
To validate the importance of each module in our proposed method, particularly the structure estimation branch (SEB) and the structure-texture fusion module (STF), we design four different network architectures on the SCI1KW dataset and conduct ablation experiments in both in-scale and out-of-scale scenarios.
In the experiments, we first explore the importance of the SEB module. To do this, we remove the SEB module from the model framework and perform feature fusion using direct upsampling of features. As shown in Table~\ref{tab3}, the use of the SEB module significantly improves performance metrics in both in-scale and out-of-scale cases, indicating that the SEB plays a crucial role in feature aggregation and information extraction. Next, we assess the effectiveness of the STF module. In this experiment, we directly combine the features from the Structure Estimation Branch and the Texture Estimation Branch before decoding. The results demonstrate that removing the STF module leads to a substantial decline in all performance metrics, highlighting its significance in detail reconstruction and visual quality enhancement.

When both the SEB and STF modules are absent, the model performs the worst, further validating the synergistic and essential roles of these two modules in our framework. These experimental results clearly indicate that our proposed modules not only effectively enhance the overall performance of the model but also better preserve details and improve visual quality when handling complex images. Meanwhile, to validate the effectiveness of B-splines, we conducted another ablation experiment, where we used the Fourier representation in addition to B-splines. The results show that using the Fourier representation leads to a significant decline in all performance metrics, which validates that B-splines are more effective than the Fourier representation for handling screen content images.

\subsection{Comparisons for Natural Images}
To further validate the performance of our model on real-world natural images, we retrain it on the widely used DIV2K dataset\cite{agustsson2017ntire}. In these experiments, we evaluate STEN on several benchmark datasets, including DIV2KW, Set5W, Set14W, B100W, and Urban100W\cite{son2021srwarp}. The results are presented in \cref{tab:tabel4}, demonstrating that STEN performs well on natural image benchmarks, despite a slight decline in performance in certain cases. These results further highlight the robustness of STEN across a wide range of datasets, including more diverse natural images, making it applicable to practical scenarios.

\begin{figure}[t]
    \centering
    \begin{minipage}[t]{0.15\textwidth}
        \centering
        \includegraphics[width=\textwidth,height=2.2cm]{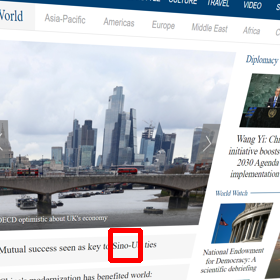}
    \end{minipage}
    \begin{minipage}[t]{0.1\textwidth}
        \centering
        \includegraphics[width=\textwidth]{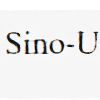}
    \end{minipage}
    \begin{minipage}[t]{0.1\textwidth}
        \centering
        \includegraphics[width=\textwidth]{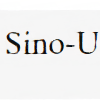}
    \end{minipage}
    \begin{minipage}[t]{0.1\textwidth}
        \centering
        \includegraphics[width=\textwidth]{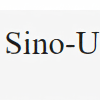}
    \end{minipage}   
    \vspace{0.5mm}

    \begin{minipage}[t]{0.15\textwidth}
        \centering
        \includegraphics[width=\textwidth,height=2.4cm]{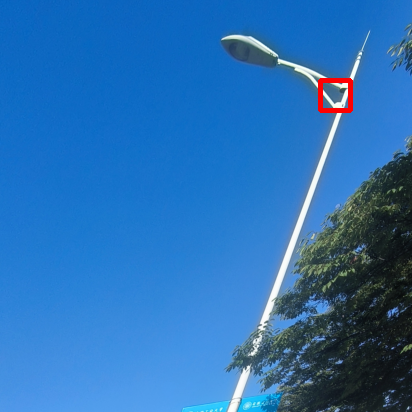}
        \vspace{\baselineskip}
        \centerline{Warped Image}
    \end{minipage}
    \begin{minipage}[t]{0.1\textwidth}
        \centering
        \includegraphics[width=\textwidth]{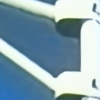}
        \vspace{\baselineskip}
        \centerline{SRWarp\cite{son2021srwarp}}
    \end{minipage}
    \begin{minipage}[t]{0.1\textwidth}
        \centering
        \includegraphics[width=\textwidth]{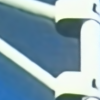}
        \vspace{\baselineskip}
        \centerline{LTEW \cite{lee2022learning}}
    \end{minipage}
    \begin{minipage}[t]{0.1\textwidth}
        \centering
        \includegraphics[width=\textwidth]{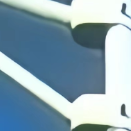}
        \vspace{\baselineskip}
        \centerline{STEN(Ours)} 
    \end{minipage}   
    \vspace{-1em}
    \caption{Quantitative comparison on real-world images.}
    \label{fig7}
    \vspace{-1em}
\end{figure}

\subsection{Comparisons for the Real-world Images}
We conducted experiments on real-world images to validate the effectiveness of our proposed error estimation model. In the distorted images captured using a camera, the transformation matrix parameters were unknown at the time of capture. Subsequently, we applied the trained error estimation model to estimate these unknown parameters.
Due to the lack of ground truth images (GT), we were unable to perform quantitative analysis; therefore, we opted for a qualitative comparison between STEN and SRWarp\cite{son2021srwarp} as well as LTEW\cite{lee2022learning}. As shown in Fig.~\ref{fig7}, we made predictions for both screen content images and natural images. The results indicate that our error estimation model can effectively predict the parameters of the transformation matrix, thereby validating the effectiveness of the proposed method.

\section{Conclusion}
In this work, we propose an innovative structure-texture enhancement dual-branch model, named STEN, to achieve arbitrary scale super-resolution (SR) and homography unwarping. Our STEN model includes a structure estimation branch and a texture estimation branch. The texture estimation branch uses the implicit B-spline neural representation of transformed images to handle coordinate transformation and improve the synthesized texture details.
At the same time, the structure estimation branch enhances structural features by using global transformation blocks and local transformation blocks, facilitating effective global modeling and thereby better representing the overall shape and structure of the image.
Additionally, our STEN model also introduces the transformation self-estimation module to process unknown transformation cases in real-world scenarios. Experiments demonstrate that our proposed STEN method significantly outperforms existing techniques in homography unwarping tasks across three publicly available SCI datasets, five natural image datasets, and the real-world images.
{\small
\bibliographystyle{cvm}
\bibliography{cvmbib}

\begin{thebibliography}{10}\itemsep=-1pt

\bibitem{agoston2005computer}
M.~K. Agoston and M.~K. Agoston.
\newblock {\em Computer graphics and geometric modeling}, volume~1.
\newblock Springer, 2005.

\bibitem{agustsson2017ntire}
E.~Agustsson and R.~Timofte.
\newblock Ntire 2017 challenge on single image super-resolution: Dataset and study.
\newblock In {\em Proceedings of the IEEE conference on computer vision and pattern recognition workshops}, pages 126--135, 2017.

\bibitem{aleman2014automatic}
M.~Alem{\'a}n-Flores, L.~Alvarez, L.~Gomez, and D.~Santana-Cedr{\'e}s.
\newblock Automatic lens distortion correction using one-parameter division models.
\newblock {\em Image Processing On Line}, 4:327--343, 2014.

\bibitem{ashish2017attention}
V.~Ashish.
\newblock Attention is all you need.
\newblock {\em Advances in neural information processing systems}, 30:I, 2017.

\bibitem{bradski2000opencv}
G.~Bradski.
\newblock The opencv library.
\newblock {\em Dr. Dobb’s Journal of Software Tools}, 2000.

\bibitem{chen2017tracking}
D.~Chen, Y.~Zhang, and S.~Li.
\newblock Tracking control of robot manipulators with unknown models: A jacobian-matrix-adaption method.
\newblock {\em IEEE Transactions on Industrial Informatics}, 14(7):3044--3053, 2017.

\bibitem{chen2021learning}
Y.~Chen, S.~Liu, and X.~Wang.
\newblock Learning continuous image representation with local implicit image function.
\newblock In {\em Proceedings of the IEEE/CVF conference on computer vision and pattern recognition}, pages 8628--8638, 2021.

\bibitem{chiang1996efficient}
M.-C. Chiang and T.~E. Boult.
\newblock Efficient image warping and super-resolution.
\newblock In {\em Proceedings Third IEEE Workshop on Applications of Computer Vision. WACV'96}, pages 56--61. IEEE, 1996.

\bibitem{chiang2000efficient}
M.-C. Chiang and T.~E. Boult.
\newblock Efficient super-resolution via image warping.
\newblock {\em Image and Vision Computing}, 18(10):761--771, 2000.

\bibitem{chitra2021retracted}
K.~Chitra and C.~Vennila.
\newblock Retracted article: A novel patch selection technique in ann b-spline bayesian hyperprior interpolation vlsi architecture using fuzzy logic for highspeed satellite image processing.
\newblock {\em Journal of Ambient Intelligence and Humanized Computing}, 12(6):6491--6504, 2021.

\bibitem{cornillere2019blind}
V.~Cornillere, A.~Djelouah, W.~Yifan, O.~Sorkine-Hornung, and C.~Schroers.
\newblock Blind image super-resolution with spatially variant degradations.
\newblock {\em ACM Transactions on Graphics (TOG)}, 38(6):1--13, 2019.

\bibitem{dai2019second}
T.~Dai, J.~Cai, Y.~Zhang, S.-T. Xia, and L.~Zhang.
\newblock Second-order attention network for single image super-resolution.
\newblock In {\em Proceedings of the IEEE/CVF conference on computer vision and pattern recognition}, pages 11065--11074, 2019.

\bibitem{duffin1952class}
R.~J. Duffin and A.~C. Schaeffer.
\newblock A class of nonharmonic fourier series.
\newblock {\em Transactions of the American Mathematical Society}, 72(2):341--366, 1952.

\bibitem{foley1996computer}
J.~D. Foley.
\newblock {\em Computer graphics: principles and practice}, volume 12110.
\newblock Addison-Wesley Professional, 1996.

\bibitem{gottlieb1997gibbs}
D.~Gottlieb and C.-W. Shu.
\newblock On the gibbs phenomenon and its resolution.
\newblock {\em SIAM review}, 39(4):644--668, 1997.

\bibitem{gottlieb2011review}
S.~Gottlieb, J.-H. Jung, and S.~Kim.
\newblock A review of david gottlieb’s work on the resolution of the gibbs phenomenon.
\newblock {\em Communications in Computational Physics}, 9(3):497--519, 2011.

\bibitem{hu2019meta}
X.~Hu, H.~Mu, X.~Zhang, Z.~Wang, T.~Tan, and J.~Sun.
\newblock Meta-sr: A magnification-arbitrary network for super-resolution.
\newblock In {\em Proceedings of the IEEE/CVF conference on computer vision and pattern recognition}, pages 1575--1584, 2019.

\bibitem{kingma2014adam}
D.~P. Kingma and J.~Ba.
\newblock Adam: A method for stochastic optimization.
\newblock {\em arXiv preprint arXiv:1412.6980}, 2014.

\bibitem{lee2022learning}
J.~Lee, K.~P. Choi, and K.~H. Jin.
\newblock Learning local implicit fourier representation for image warping.
\newblock In {\em European Conference on Computer Vision}, pages 182--200. Springer, 2022.

\bibitem{lee2022local}
J.~Lee and K.~H. Jin.
\newblock Local texture estimator for implicit representation function.
\newblock In {\em Proceedings of the IEEE/CVF conference on computer vision and pattern recognition}, pages 1929--1938, 2022.

\bibitem{lehmann2001addendum}
T.~M. Lehmann, C.~Gonner, and K.~Spitzer.
\newblock Addendum: B-spline interpolation in medical image processing.
\newblock {\em IEEE transactions on medical imaging}, 20(7):660--665, 2001.

\bibitem{li2020application}
L.~Li.
\newblock Application of cubic b-spline curve in computer-aided animation design.
\newblock {\em Computer-Aided Design and Applications}, 18(S1):43--52, 2020.

\bibitem{li2017robust}
L.~Li, W.~Liu, and W.~Xing.
\newblock Robust radial distortion correction from a single image.
\newblock In {\em 2017 IEEE 15th Intl Conf on Dependable, Autonomic and Secure Computing, 15th Intl Conf on Pervasive Intelligence and Computing, 3rd Intl Conf on Big Data Intelligence and Computing and Cyber Science and Technology Congress (DASC/PiCom/DataCom/CyberSciTech)}, pages 766--772. IEEE, 2017.

\bibitem{liang2021swinir}
J.~Liang, J.~Cao, G.~Sun, K.~Zhang, L.~Van~Gool, and R.~Timofte.
\newblock Swinir: Image restoration using swin transformer.
\newblock In {\em Proceedings of the IEEE/CVF international conference on computer vision}, pages 1833--1844, 2021.

\bibitem{lim2017enhanced}
B.~Lim, S.~Son, H.~Kim, S.~Nah, and K.~Mu~Lee.
\newblock Enhanced deep residual networks for single image super-resolution.
\newblock In {\em Proceedings of the IEEE conference on computer vision and pattern recognition workshops}, pages 136--144, 2017.

\bibitem{mok2020large}
T.~C. Mok and A.~C. Chung.
\newblock Large deformation diffeomorphic image registration with laplacian pyramid networks.
\newblock In {\em Medical Image Computing and Computer Assisted Intervention--MICCAI 2020: 23rd International Conference, Lima, Peru, October 4--8, 2020, Proceedings, Part III 23}, pages 211--221. Springer, 2020.

\bibitem{pak2023b}
B.~Pak, J.~Lee, and K.~H. Jin.
\newblock B-spline texture coefficients estimator for screen content image super-resolution.
\newblock In {\em Proceedings of the IEEE/CVF Conference on Computer Vision and Pattern Recognition}, pages 10062--10071, 2023.

\bibitem{prasad2022nurbs}
A.~D. Prasad, A.~Balu, H.~Shah, S.~Sarkar, C.~Hegde, and A.~Krishnamurthy.
\newblock Nurbs-diff: A differentiable programming module for nurbs.
\newblock {\em Computer-Aided Design}, 146:103199, 2022.

\bibitem{santana2016iterative}
D.~Santana-Cedr{\'e}s, L.~Gomez, M.~Alem{\'a}n-Flores, A.~Salgado, J.~Esclar{\'\i}n, L.~Mazorra, and L.~Alvarez.
\newblock An iterative optimization algorithm for lens distortion correction using two-parameter models.
\newblock {\em Image Processing On Line}, 6:326--364, 2016.

\bibitem{sarlin2021back}
P.-E. Sarlin, A.~Unagar, M.~Larsson, H.~Germain, C.~Toft, V.~Larsson, M.~Pollefeys, V.~Lepetit, L.~Hammarstrand, F.~Kahl, et~al.
\newblock Back to the feature: Learning robust camera localization from pixels to pose.
\newblock In {\em Proceedings of the IEEE/CVF conference on computer vision and pattern recognition}, pages 3247--3257, 2021.

\bibitem{schoenberg1946contributions}
I.~J. Schoenberg.
\newblock Contributions to the problem of approximation of equidistant data by analytic functions. part b. on the problem of osculatory interpolation. a second class of analytic approximation formulae.
\newblock {\em Quarterly of Applied Mathematics}, 4(2):112--141, 1946.

\bibitem{sitzmann2020implicit}
V.~Sitzmann, J.~Martel, A.~Bergman, D.~Lindell, and G.~Wetzstein.
\newblock Implicit neural representations with periodic activation functions.
\newblock {\em Advances in neural information processing systems}, 33:7462--7473, 2020.

\bibitem{son2021srwarp}
S.~Son and K.~M. Lee.
\newblock Srwarp: Generalized image super-resolution under arbitrary transformation.
\newblock In {\em Proceedings of the IEEE/CVF conference on computer vision and pattern recognition}, pages 7782--7791, 2021.

\bibitem{umetani1989resolved}
Y.~Umetani, K.~Yoshida, et~al.
\newblock Resolved motion rate control of space manipulators with generalized jacobian matrix.
\newblock {\em IEEE Transactions on robotics and automation}, 5(3):303--314, 1989.

\bibitem{unser1993b}
M.~Unser, A.~Aldroubi, and M.~Eden.
\newblock B-spline signal processing. ii. efficiency design and applications.
\newblock {\em IEEE transactions on signal processing}, 41(2):834--848, 1993.

\bibitem{unser1993b1}
M.~Unser, A.~Aldroubi, and M.~Eden.
\newblock B-spline signal processing. ii. efficiency design and applications.
\newblock {\em IEEE transactions on signal processing}, 41(2):834--848, 1993.

\bibitem{van2014scikit}
S.~Van~der Walt, J.~L. Sch{\"o}nberger, J.~Nunez-Iglesias, F.~Boulogne, J.~D. Warner, N.~Yager, E.~Gouillart, and T.~Yu.
\newblock scikit-image: image processing in python.
\newblock {\em PeerJ}, 2:e453, 2014.

\bibitem{von2020lm}
L.~Von~Stumberg, P.~Wenzel, N.~Yang, and D.~Cremers.
\newblock Lm-reloc: Levenberg-marquardt based direct visual relocalization.
\newblock In {\em 2020 International Conference on 3D Vision (3DV)}, pages 968--977. IEEE, 2020.

\bibitem{voulodimos2018deep}
A.~Voulodimos, N.~Doulamis, A.~Doulamis, and E.~Protopapadakis.
\newblock Deep learning for computer vision: A brief review.
\newblock {\em Computational intelligence and neuroscience}, 2018(1):7068349, 2018.

\bibitem{wang2018esrgan}
X.~Wang, K.~Yu, S.~Wu, J.~Gu, Y.~Liu, C.~Dong, Y.~Qiao, and C.~Change~Loy.
\newblock Esrgan: Enhanced super-resolution generative adversarial networks.
\newblock In {\em Proceedings of the European conference on computer vision (ECCV) workshops}, pages 0--0, 2018.

\bibitem{xiao2024towards}
J.~Xiao, Z.~Lyu, C.~Zhang, Y.~Ju, C.~Shui, and K.-M. Lam.
\newblock Towards progressive multi-frequency representation for image warping.
\newblock In {\em Proceedings of the IEEE/CVF Conference on Computer Vision and Pattern Recognition}, pages 2995--3004, 2024.

\bibitem{yang2021implicit}
J.~Yang, S.~Shen, H.~Yue, and K.~Li.
\newblock Implicit transformer network for screen content image continuous super-resolution.
\newblock {\em Advances in Neural Information Processing Systems}, 34:13304--13315, 2021.

\bibitem{yang2019deep}
W.~Yang, X.~Zhang, Y.~Tian, W.~Wang, J.-H. Xue, and Q.~Liao.
\newblock Deep learning for single image super-resolution: A brief review.
\newblock {\em IEEE Transactions on Multimedia}, 21(12):3106--3121, 2019.

\bibitem{zhang2018residual}
Y.~Zhang, Y.~Tian, Y.~Kong, B.~Zhong, and Y.~Fu.
\newblock Residual dense network for image super-resolution.
\newblock In {\em Proceedings of the IEEE conference on computer vision and pattern recognition}, pages 2472--2481, 2018.

\bibitem{zheng2024efficient}
L.~Zheng, J.~Zhu, J.~Shi, and S.~Weng.
\newblock Efficient mixed transformer for single image super-resolution.
\newblock {\em Engineering Applications of Artificial Intelligence}, 133:108035, 2024.

\bibitem{zhou2005large}
K.~Zhou, J.~Huang, J.~Snyder, X.~Liu, H.~Bao, B.~Guo, and H.-Y. Shum.
\newblock Large mesh deformation using the volumetric graph laplacian.
\newblock {\em ACM Transactions on Graphics}, 24(3):496--503, 2005.

\bibitem{zhu2017unpaired}
J.-Y. Zhu, T.~Park, P.~Isola, and A.~A. Efros.
\newblock Unpaired image-to-image translation using cycle-consistent adversarial networks.
\newblock In {\em Proceedings of the IEEE international conference on computer vision}, pages 2223--2232, 2017.

\end{thebibliography}
}

\end{document}